\pdfoutput=1

\documentclass[11pt]{article}

\usepackage[]{ACL2023}

\usepackage{times}
\usepackage{latexsym}

\usepackage[T1]{fontenc}

\usepackage[utf8]{inputenc}

\usepackage{microtype}

\usepackage{inconsolata}

\usepackage{kotex}
\usepackage{graphicx}
\usepackage{graphics}
\usepackage{xspace}
\usepackage{amsfonts}
\usepackage{amsmath}
\usepackage{algorithm}
\usepackage{algpseudocode}
\usepackage{arydshln}
\usepackage{subcaption}
\usepackage{multirow}
\usepackage{booktabs}
\usepackage{pgfplots}
\usepackage{caption}
\usepackage{verbatim}
\usepackage{tcolorbox}
\usepackage{xcolor}

\title{Deep Exploration of Cross-Lingual Zero-Shot Generalization in

Instruction Tuning}

\author{Janghoon Han{\textsuperscript{}\thanks{\;\;\;indicates equal contribution.}} \quad Changho Lee{\textsuperscript{$\ast$}} \quad Joongbo Shin{\textsuperscript{}}\\ {\bf Stanley Jungkyu Choi{\textsuperscript{}} \quad Honglak Lee{\textsuperscript{}} \quad Kyunghoon Bae{\textsuperscript{}}} \\
        {\textsuperscript{}}LG AI Research\\ 
        \texttt{\{janghoon.han,changho.lee\}@lgresearch.ai} 
        \\
        }

\begin{document}
\maketitle
\begin{abstract}
Instruction tuning has emerged as a powerful technique, significantly boosting zero-shot performance on unseen tasks. While recent work has explored cross-lingual generalization by applying instruction tuning to multilingual models, previous studies have primarily focused on English, with a limited exploration of non-English tasks.
For an in-depth exploration of cross-lingual generalization in instruction tuning, we perform instruction tuning individually for two distinct language meta-datasets. Subsequently, we assess the performance on unseen tasks in a language different from the one used for training. To facilitate this investigation, we introduce a novel non-English meta-dataset named "KORANI" (Korean Natural Instruction), comprising 51 Korean benchmarks. Moreover, we design cross-lingual templates to mitigate discrepancies in language and instruction-format of the template between training and inference within the cross-lingual setting.
Our experiments reveal consistent improvements through cross-lingual generalization in both English and Korean, outperforming baseline by average scores of 20.7\% and 13.6\%, respectively. Remarkably, these enhancements are comparable to those achieved by monolingual instruction tuning and even surpass them in some tasks. The result underscores the significance of relevant data acquisition across languages over linguistic congruence with unseen tasks during instruction tuning\footnote{\url{https://github.com/CHLee0801/KORANI-Instruction-Tuning}}.

\end{abstract}

\section{Introduction}


Recent studies have highlighted Instruction Tuning, where large language models are fine-tuned using instructions (templates) for various tasks, resulting in significant improvements in zero-shot performance on unseen tasks \cite{Flan,T0,supernatural,Flan_t5,seonghyeon, ouyang,instruction_tuning_zhong}. 
Several works have investigated the effectiveness of instruction tuning to show cross-lingual zero-shot generalization.
For example, \citet{supernatural,bloomz,joel,bactarian} apply instruction tuning to multilingual large language models \cite{mbert, mt5,bloom,xglm}, pre-trained with multiple languages, and demonstrate cross-lingual generalization ability by achieving meaningful performance enhancements for unseen tasks in other languages.

However, prior investigations into cross-lingual generalization through instruction tuning, as observed in works by \citet{supernatural} and \citet{bloomz}, have predominantly focused on English and have limited diversity of tasks in non-English.
This limited scope makes it challenging to thoroughly evaluate the effectiveness of cross-lingual zero-shot generalization, as direct comparisons with monolingual instruction tuning within the same language are not attainable. 
Moreover, previous research lacks validation of cross-lingual generalization of instruction tuning for languages other than English, and the evaluation of non-English datasets has been confined to specific tasks, offering only partial understanding.

To fill these gaps, our study undertakes a comprehensive investigation of the cross-lingual zero-shot generalization in instruction tuning. We define a new cross-lingual setting as the case where the training and inference language differs. In the setting, we instruction tune for two languages meta-dataset separately and evaluate the other language's unseen tasks. Specifically, to examine how effective cross-lingual zero-shot generalization is for a different language, we conduct a comparative analysis between cross-lingual instruction tuning and monolingual instruction tuning, where the latter indicates models trained and tested on the same language's meta-dataset.

The collection of a non-English meta-dataset \cite{meta-dataset} is imperative for the comprehensive examination of cross-lingual generalization of instruction tuning. 
However, collecting diverse supervised task datasets for non-English poses a substantial challenge due to the limited availability of open-source data in non-English languages compared to English. To address this issue, we propose a novel non-English language meta-dataset named \textbf{KORANI}, short for \textbf{KOR}e\textbf{A}n \textbf{N}atural \textbf{I}nstruction. This meta-dataset comprises 51 diverse Korean benchmarks, including 34 NLU benchmarks and 17 NLG benchmarks. Notably, KORANI surpasses the quantity of non-English benchmarks explored in previous multilingual research \cite{supernatural,bloomz} and approaches the size of P3 datasets \cite{T0}, which we employ as English benchmarks in our study.

Furthermore, in the cross-lingual setting, the language and instructional format of templates are different during the training and inference phase. These discrepancies in the template may contribute to suboptimal model performance \cite{bloomz,consistency,ins_robust}.
To address this issue, we construct cross-lingual templates to align the template between the training and inference phases.

Our experiments show that cross-lingual instruction tuning consistently improves the zero-shot performance of unseen tasks in both English and Korean. Surprisingly, these cross-lingual performances are comparable to those of monolingual instruction-tuned models across various tasks, and they even surpass some tasks when cross-lingual templates are applied. These findings suggest that learning relevant tasks, even in different languages, is more crucial for performance improvement than ensuring linguistic congruence within unseen tasks in instruction tuning. Furthermore, our findings reframe the traditional view that cross-lingual instruction tuning merely enhances performance in low-resource languages, suggesting that it could serve as a viable alternative to monolingual instruction tuning.

Our contributions are summarized as follows:
\begin{itemize}
 \item Our study enhances the understanding of cross-lingual instruction tuning by demonstrating that it can match the performance of monolingual tuning, emphasizing the importance of learning relevant tasks across languages.
 \item We introduce a new dataset called \textbf{KORANI}, comprising diverse Korean benchmarks, which provides a valuable resource for instruction tuning in non-English languages.
 \item We introduce cross-lingual templates to both the P3 dataset and KORANI, and confirm the robustness of cross-lingual zero-shot generalization achieved with these templates.
\end{itemize}


\begin{figure*}[ht!]
\centering
    \includegraphics[width=0.9\textwidth]{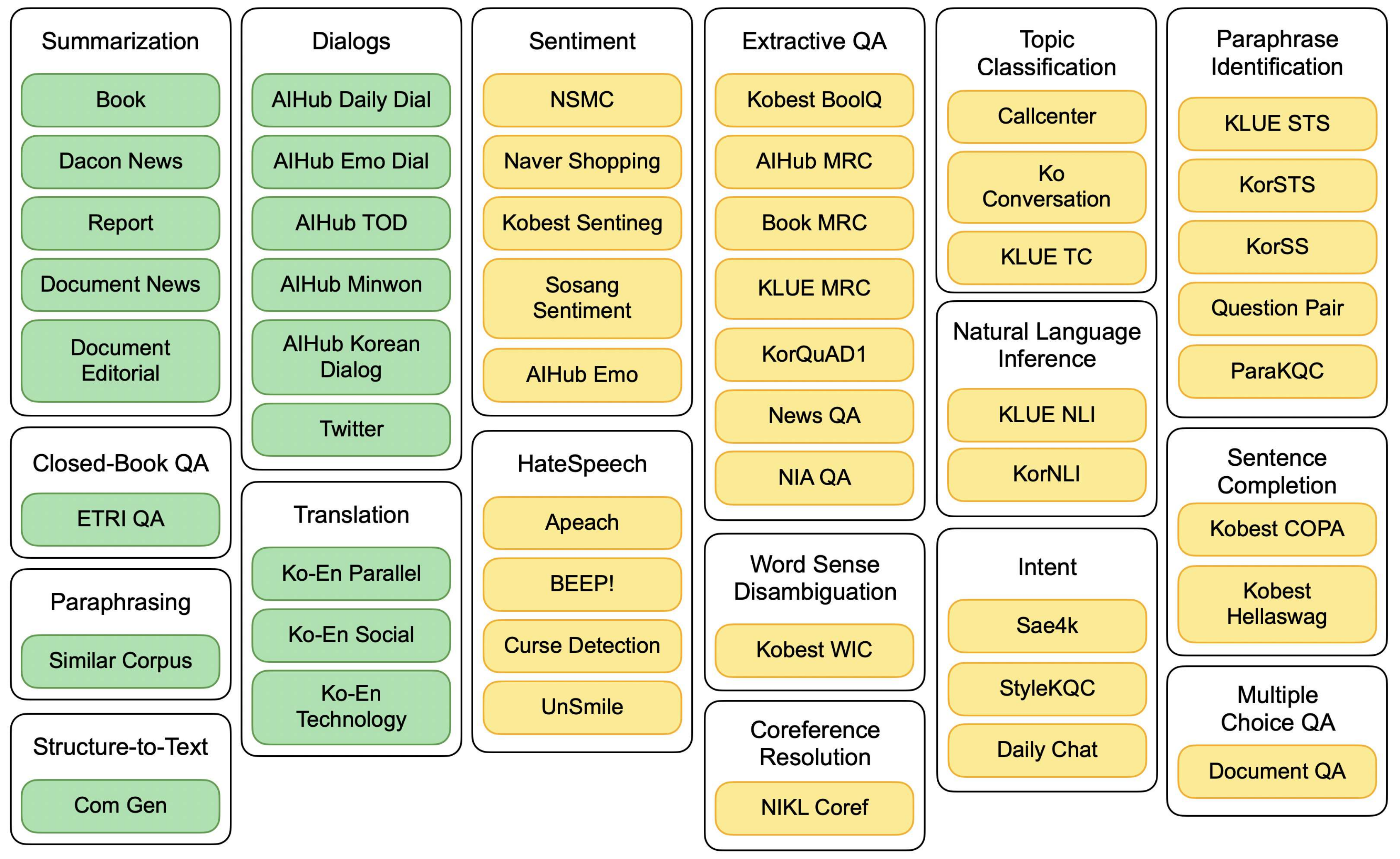}
\caption{\textsc{KORANI} datasets and task taxonomy. Green datasets are NLG datasets. Yellow datasets are NLU datasets. We follow task categorization from \citet{T0}}
\label{fig:KORANI Datasets}
\vspace{-5mm}
\end{figure*}

\section{Related Work}
\subsection{Instruction Tuning}
Instruction Tuning represents a learning methodology that enhances the zero-shot performance of unseen tasks by leveraging various Natural Language Processing (NLP) tasks. 
Instruction tuning explicitly trains NLP tasks using a multi-task training approach, and leverages templates to learn the salient characteristics of these tasks. By combining datasets and templates, instruction tuning induces robust generalization for unseen tasks with new templates adapted to assist the model for problem solving capability.
Previous studies \cite{Flan,supernatural} defined task clusters in various ways, and we follow the T0 \cite{T0} task taxonomy. T0 leverages the templates source software application \cite{promptsource} to collect English templates, which are subsequently used to form a Public Pool of Prompts (P3) for learning. We employ P3 for English meta-dataset which comprises 12 tasks and 62 datasets.

\subsection{Cross-lingual Task Generalization in Instruction Tuning}
Previous studies \cite{supernatural, bloomz} have contributed to the understanding of cross-lingual zero-shot generalization within the instruction tuning. \citet{supernatural} first extend the boundaries by introducing both English and multilingual models trained on instruction formatted datasets. Their study construct a meta-dataset encompassing 76 task types and 1616 datasets, which included 576 datasets across 54 non-English languages. 
\citet{bloomz} investigate the efficacy of English-only instruction tuning in enhancing performance on non-English held-out tasks. 
Moreover, they introduce meta-datasets xP3 and xP3\_mt, enriched with multilingual datasets and machine-translated templates, demonstrating further improved zero-shot performance in both English and non-English tasks.

However, the studies by \citet{supernatural, bloomz} have some limitations. First, the majority of the training data is composed of English, with only a minor portion in non-English languages. This setup confirms the transfer effectiveness from English to other languages but fails to thoroughly explore the transfer capabilities from non-English languages to other languages \cite{limitation_2_2,limitation_2_3,limitation_2_4,ouyang}. Second, although the evaluations include a variety of non-English languages, the number of tasks per language is limited, hindering a comprehensive validation across diverse task types. Most importantly, while these studies verify the performance enhancements due to cross-lingual transfer, they do not address how these improvements compare to those achieved through monolingual instruction tuning.

To address this gap, we conduct a more comprehensive study to investigate cross-lingual zero-shot generalization in instruction tuning. Specifically, we construct and learn meta-datasets in both English and Korean and demonstrate the cross-lingual zero-shot generalization efficacy by directly comparing the performance of instruction tuning trained in other languages to that in the same language.


\section{Measuring Cross-lingual Zero-shot Generalization}

To investigate the effect of cross-lingual zero-shot generalization in instruction tuning, we perform instruction tuning independently for both English and Korean. We then measure the cross-lingual generalization by evaluating the models' performance on unseen tasks in the other language, respectively. 

\begin{figure*}[ht!]
\centering
    \includegraphics[width=\textwidth]{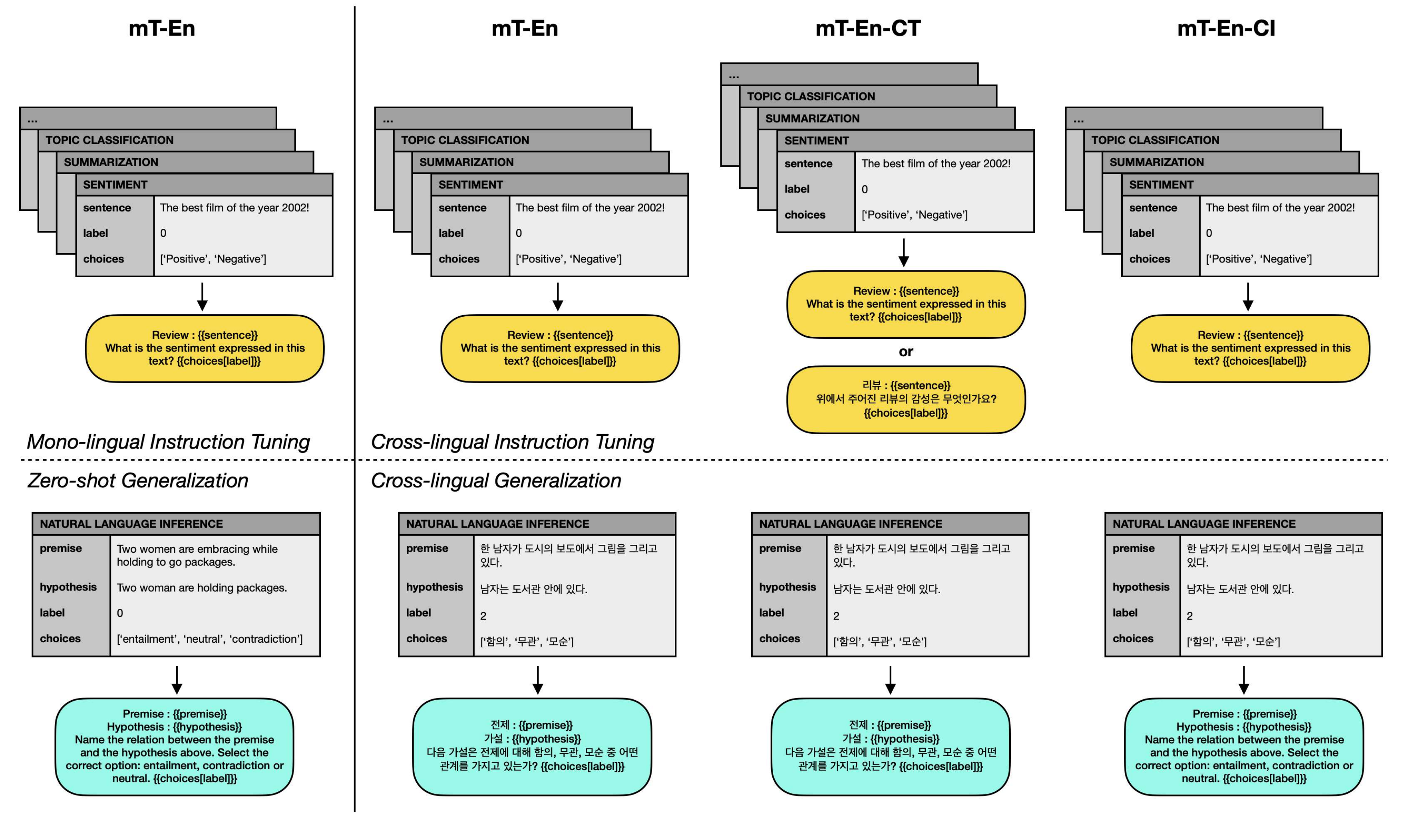}
\caption{Comparison of model variants mT-En, mT-En-CT, and mT-En-CI on samples from Rotten Tomatoes, esNLI for P3 \cite{T0}, and KLUE NLI for KORANI. The dashed line differentiates training and evaluation, while the solid line distinguishes monolingual and cross-lingual generalization. mT-En-CT pairs English datasets with either English or Korean templates during training, and mT-En-CI pairs Korean datasets with English templates during evaluation.}
\label{fig:templates_diagram}
\vspace{-5mm}
\end{figure*}

\subsection{Dataset for Instruction Tuning}
\subsubsection{KORANI: KOReAn Natural Instructions}
For the Korean instruction tuning, we introduce a novel meta-dataset named KORANI. KORANI is the first collection of various Korean NLP tasks available in the Korean research community, which then transformed into an instructional format that describes the task in plain language. The significance of our research lies in the fact that, unlike previous studies \cite{bloomz,bactarian} that relied on machine translation, we curate and generate high-quality datasets through meticulous human effort by experts. The collection process comprises benchmark collection, instruction creation, and quality control.

\paragraph{Benchmark Collection}
Creating an instruction tuning dataset with numerous different tasks from scratch can be a resource-intensive process. To overcome this challenge, we collected 51 existing Korean benchmarks from various eminent sources such as 
AIHub\footnote{\href{https://www.aihub.or.kr}{https://www.aihub.or.kr}}, Korpora\footnote{\href{https://ko-nlp.github.io/Korpora/en-docs}{https://ko-nlp.github.io/Korpora/en-docs}}, Github, Huggingface, KLUE\footnote{\href{https://klue-benchmark.com}{https://klue-benchmark.com}} \cite{klue}, Korquad\footnote{\href{https://korquad.github.io}{https://korquad.github.io}}, and ETRI\footnote{\href{https://nanum.etri.re.kr}{https://nanum.etri.re.kr}} 
including both language understanding and language generation tasks. Then, collected datasets are refined and categorized into task clusters. Some of KORANI datasets had no explicitly defined task. For those datasets, we define task and transform the dataset into an organized form based on careful consideration of the dataset's purpose and available data-labels included in the dataset. KORANI consists of 17 task clusters using heuristic rules proposed by \citet{T0} as illustrated in Figure \ref{fig:KORANI Datasets}. Please refer to Appendix \ref{sec:raw data transformation} for more details about dataset collection.

\paragraph{Instruction Creation}
For each benchmark dataset, we manually manufacture 10 natural language instructions.
We collaborated with 10 experienced NLP experts to create qualitative templates. Contributors were provided a detail guide to ensure that they utilize a various data-labels in the dataset. To create diverse and well-refined templates, we encouraged contributors to be open to their own style while providing strict guidelines for grammatical accuracy and clarity in natural language instructions. 

\paragraph{Quality Control}
For quality control, we removed duplicate instances, and adjusted label imbalances in each task. We went through a peer-review session with two other expert contributors per dataset five times at minimum to ensure the quality of templates. The extensive cross-validation was iteratively performed until reviewers were content with the quality of the datasets.

\subsubsection{English Instruction Tuning Benchmarks}
We utilized a Public Pool of Prompts (P3) following \citet{T0} as the English meta-dataset. Our experiment involves 62 datasets and templates for each task provided by \citet{promptsource}. We split the datasets into 12 task clusters shown in Appendix \ref{sec:p3_figure}. 

\subsubsection{Statistics of KORANI and P3}
Table \ref{table:KORANI Statistics} shows various statistics and comparisons between the KORANI and P3. KORANI consists of 51 tasks divided into 17 clusters, each comprising a comparable number of Natural Language Generation (NLG) and Natural Language Understanding (NLU) tasks. In contrast, the P3 is more heavily focused on NLU tasks. 

\begin{table}[]
\centering
    \resizebox{\columnwidth}{!}{
\begin{tabular}{ccc}
    \toprule
       Statistics & \textsc{KORANI} & \textsc{P3} \\
    \midrule
        \# of datasets & 51 & 62 \\
        \# of NLU datasets & 34 & 51 \\
        \# of NLG datasets & 17 & 11 \\
        \# of tasks & 17 & 12 \\
        avg. \# of templates (per dataset) & 10 & 8.02 \\
        avg. \# of cross-lingual templates (per dataset) & 3.76 & 3.45 \\
        avg. \# of instances (per dataset) & 4,768 & 4,388 \\
        avg. \# of input tokens (per dataset) & 179.8 & 187.64 \\
        avg. \# of output tokens (per dataset) & 16.35 & 11.06 \\
    \bottomrule
\end{tabular}}
\caption{Statistics of \textsc{KORANI} and P3. Training instances per dataset are limited to 5k maximum.}
\label{table:KORANI Statistics}
\vspace{-5mm}
\end{table} 

\subsection{Addressing Templates Misalignment Challenges in Cross-Lingual Instruction Tuning Scenarios}
To evaluate the cross-lingual zero-shot generalization, we conduct instruction tuning in one language and assess the model's performance on unseen tasks in the other language. However, since the training templates and inference templates are taken from different language datasets (KORANI and P3), it raises the possibility of performance degradation from template misalignment, which might lead to suboptimal performance \cite{T0, supernatural,bloomz,consistency,ins_robust}. The misalignment primarily occurs in two aspects. One is the linguistic misalignment of the templates \cite{bloomz}, which stems from the grammatical and semantic differences between the two languages. The other is the misalignment in the instructional format \cite{format_align1,format_align2,consistency,ins_robust}, which stems from template style differences such as ordering description in templates and level of explanation detail about the task.

We introduce cross-lingual templates to mitigate the challenges of template misalignment in cross-lingual instruction tuning scenarios. To maximize alignment between training and inference templates, we align the language and instructional format of the templates similar to targeting evaluation tasks.
We create an average of 3.76 and 3.45 cross-lingual templates each for KORANI and P3 meta datasets as shown in Table \ref{table:KORANI Statistics}. See the Appendix \ref{sec:cross-lingual templates generation} for more details on the creation process of cross-lingual templates.

We propose two approaches to integrate cross-lingual templates with cross-lingual settings. The first approach utilizes cross-lingual templates during the training phase, while the second approach employs cross-lingual templates during the inference phase to align language and instructional format in training and inference. Both strategies are meticulously designed to uphold structural similarity between templates used during training and inference. The linguistic and instructional format alignment empower the model to effectively adapt when it is presented with a new instruction for an unseen task.

\subsection{Model}
\label{sec:model}
To assess the zero-shot generalization capability of instruction tuning and its cross-lingual transferability between Korean and English, we employ mT5 \cite{mt5} models as the core model. The mT5 model is a publicly available multilingual model trained in 101 languages, including both English and Korean. The mT5 models encompass a range of sizes, from 300M to 13B parameters, and we employ 1.3B to 13B for our experiments.

We assess cross-lingual generalization by instruction tuning in one language and evaluating unseen tasks in the other language for both English and Korean. Moreover, we introduce cross-lingual templates in the training or inference phases to investigate the advantage of instruction alignment. For this scenario, we introduce the following model variants:

\begin{itemize}
    \item \textbf{mT-Ko, mT-En:} Models trained on KORANI and P3 datasets, respectively.
    \item \textbf{mT-Ko-CT, mT-En-CT}: Models trained on KORANI and P3 datasets, respectively by incorporating cross-lingual templates during training only, and inferenced with original templates. 
    \item \textbf{mT-Ko-CI, mT-En-CI}: Models trained on KORANI and P3 datasets respectively with original templates only, and inferenced with cross-lingual templates. 
\end{itemize}
The postfix \textbf{CT} and \textbf{CI} denote \textbf{C}ross-lingual instruction \textbf{T}raining, and \textbf{C}ross-lingual instruction \textbf{I}nference respectively.

\begin{figure*}[ht!]
\centering
    \includegraphics[width=\textwidth]{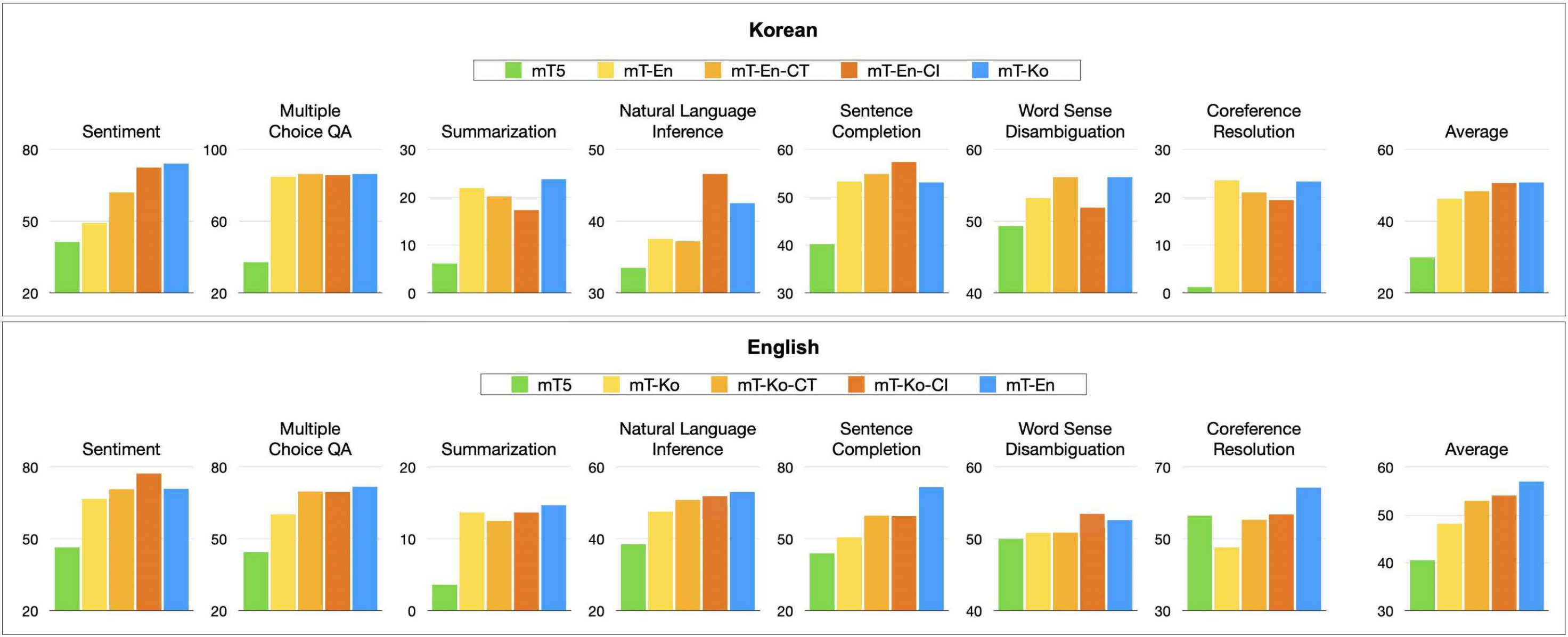}
\caption{Performance of zero-shot and cross-lingual generalization. Scores are datasets average for each task cluster. The first row denotes KORANI unseen tasks, and the second row denotes P3 unseen tasks. Average chart averages seven different task results. Appendix \ref{sec:crosslingual_breakdown} breaks down the performance by datasets.}
\label{fig:ko_instruction_tuning}
\vspace{-5mm}
\end{figure*}  

\section{Experimental Setup}

\subsection{Training}
To evaluate the zero-shot performance for various tasks, we set held-out tasks following previous studies \cite{T0,bloomz}.
In this scheme, when training a meta-dataset in a different language, we deliberately exclude tasks that correspond to the pre-defined held-out tasks in the meta-dataset. 
We have devised two distinct held-out settings. The first group, following \citet{T0}, encompasses four tasks: natural language inference, sentence completion, coreference resolution, and word sense disambiguation. In addition to the existing settings, a second group is formed to further validate the trend of cross-lingual generalization for more tasks. The second group consists of three tasks: sentiment analysis, summarization, and multiple-choice QA. 

During the training, we employ 10 distinctive templates for each dataset. For CT models, we partially replace original templates with cross-lingual templates.

We configure validation from training datasets and select the model that showed the best performance in the validation. Our experiment is in a true zero-shot setting, as we do not use any examples from held-out tasks for checkpoint selection.	

We also limit the number of examples in each dataset to 5k to avoid a skewed distribution between tasks, which end up around 180k instances per train. For more information on training, see Appendix \ref{sec:training_detail}.


\subsection{Evaluation}
We randomly sample three templates and measure the average score for each dataset. For the classification task, we employ rank classification \cite{gpt3} and for the generation task, we report ROUGE-L \cite{rouge} score for model performance, following previous work \cite{supernatural}.

The CI models follow a consistent training approach as the base models (mT-Ko and mT-En) but have a key distinction in their evaluation process by utilizing cross-lingual templates. Therefore, unlike all other models, including the CT model, the CI model gauges its performance across three randomly sampled cross-lingual templates different from the original templates. 

\begin{figure*}[ht!]
\centering
    \includegraphics[width=1\textwidth]{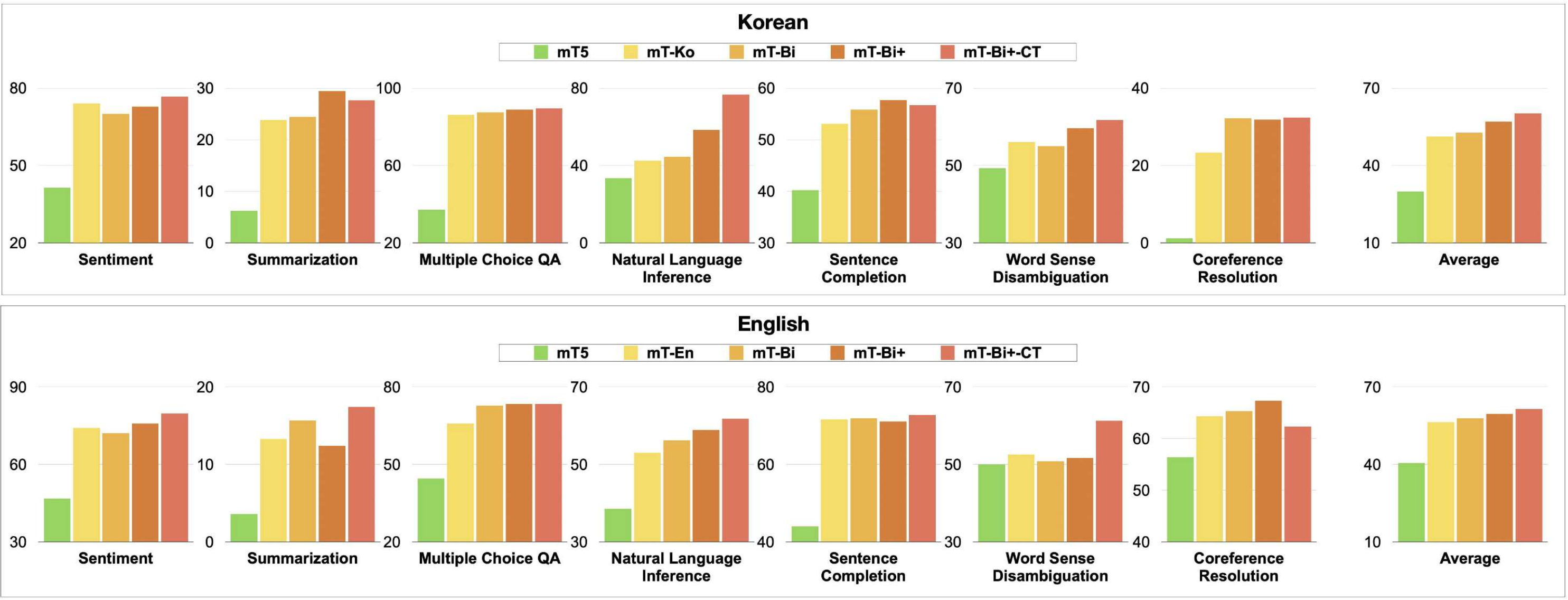}
\caption{Bilingual instruction tuning performance in KORANI and P3. mT-Bi+-CT employs the CT training method for non-target language datasets only. Appendix \ref{sec:add-bi} covers additional experiments on the cross-lingual template, and \ref{sec:bilingual_breakdown} breaks down the performance by datasets.}
\label{fig:bi-instruction-tuning}
\vspace{-5mm}
\end{figure*}  

\section{Result}
\label{sec:result}
\subsection{Cross-lingual Transfer between Korean and English}
\label{sec:cross-lingual generalization}
To assess the impact of cross-lingual zero-shot generalization, we initially conduct instruction tuning in one language and evaluate the model's performance on unseen tasks in the other language. As depicted in Figure \ref{fig:ko_instruction_tuning}, both languages exhibit notable performance enhancements even when subjected to instruction tuning conducted in a different language. Moreover, in certain tasks, the achieved performance closely resembles that of the model trained within the same language. Specifically, tasks like multiple-choice QA, summarization, and sentence completion of Korean evaluation display comparable performance between mT-En and the model mT-Ko. For English evaluation, the mT-Ko demonstrates performance akin to that of mT-En in sentiment analysis and summarization tasks.

\subsection{Effect of Cross-lingual Templates in Cross-lingual Generalization}

Furthermore, CT and CI models that incorporate well-aligned cross-lingual templates show notable performance improvements across most tasks in both languages. Specifically, in the Korean evaluation, mT-En-CT and mT-En-CI outperform mT-En. Similarly, in the English evaluation, mT-Ko-CT and mT-Ko-CI surpass mT-Ko. This finding highlights the importance of well-aligned templates in facilitating effective cross-lingual generalization.
 
The key takeaway from our experiments is that the mT-En-CI and mT-Ko-CI models, trained in a different language, achieve performance comparable to the mT-Ko and mT-En models, which are trained in the same language. This consistent trend is observed across the majority of tasks during evaluations in both Korean and English. Notably, the CI models outperform, particularly excelling in certain tasks like sentiment analysis and sentence completion. Specifically, models trained on the KORANI dataset display robust performance in sentiment analysis, while the CI models trained on the P3 dataset demonstrate exceptional sentence completion capabilities, regardless of the language evaluated. These findings indicate that training on relevant tasks in a different language can still yield significant performance, potentially even surpassing that of monolingual instruction tuning. This underscores that focusing on relevant tasks is more important than adhering to linguistic congruence for unseen tasks. For a more detailed analysis, please refer to the Appendix \ref{sec:cross_detail}.


\section{Further Analysis}
\subsection{Bilingual Instruction Tuning}
\label{sec:bilingual}
Instruction tuning in a single language sufficiently shows cross-lingual zero-shot generalization. To delve deeper into the potential synergistic effects arising from the utilization of two languages in instruction tuning together, we introduce bilingual instruction tuning. This approach combines and jointly trains two meta-datasets, KORANI and P3, then compares the performance of unseen tasks with single language instruction tuning. This experiment involves the following additional model variants:
\begin{itemize}
    \item \textbf{mT-Bi} models are trained on both the KORANI and P3 datasets using bilingual instruction tuning.
    \item \textbf{mT-Bi+} maintains the settings of mT-Bi but includes held-out tasks from a non-evaluating language.
    \item \textbf{mT-Bi+-CT} employs the same dataset composition as mT-Bi+ and further incorporates cross-lingual templates from the non-evaluating language's meta-dataset during the training phase.
\end{itemize}

Figure \ref{fig:bi-instruction-tuning} shows that training the model on a mixture of two meta-datasets results in improved performance compared to single meta-dataset training for both languages. We speculate that similar to the trend in monolingual instruction tuning \cite{Flan, T0}, where increased task diversity enhances performance, learning a broader range of tasks irrespective of language has also led to performance improvements in cross-lingual instruction tuning.

Furthermore, mT-Bi+, the model trained by adding datasets corresponding to held-out tasks in the other language, demonstrates improved performance. This result underscores that incorporating aligned datasets can guide the model to more explicitly learn the targeted unseen tasks, thereby enhancing cross-task generalization.

Lastly, when considering the mT-Bi+-CT model, which integrates cross-lingual templates into the mT-Bi+ during the training phase, consistent performance enhancements are observed for both languages. This trend aligns with Section \ref{sec:cross-lingual generalization}, emphasizing alignment of instructions facilitates improved adaptation of the model to unseen tasks.

\subsection{Template alignment: Linguistic Or Instructional Format}

In this section, we analyze whether the performance improvements in cross-lingual transfer through instruction alignment originate from linguistic factors or formatting. To do this, we employ mT-Ko and two of its variants and evaluate their performance on the held-out tasks from the P3 benchmark. The first variant, mT-Ko-Trans, merely employs a translated version of P3 templates into Korean during the inference phase. This variant aligns linguistic aspects of templates between training and inference. The second variant, mT-Ko-CI, as mentioned in Section \ref{sec:model}, encompasses instruction alignment that considers both linguistic and instructional format aspects during inference. Further illustrative examples are available in the Appendix \ref{sec:examples_of_ci_trans}.

Table \ref{table:CI-Trans} shows that mT-Ko-Trans demonstrates consistently improved performance compared to the mT-Ko evaluated using P3 templates, across most of the tasks. This observation proves that notable performance enhancement is achievable through linguistic alignment alone, as it guides the model to better adapt to new unseen templates.
When comparing the performance of mT-Ko-Trans with that of mT-Ko-CI, it becomes evident that the latter achieves higher performance. This result is attributed to the fact that while both approaches entail linguistic alignment between the training and evaluation templates, mT-Ko-CI additionally gains cross-task generalization through the alignment of structural formatting between training and evaluation templates. The result highlights that both linguistic and instructional format alignment is important in cross-lingual generalization.

\begin{table}[t]
\centering
    \resizebox{\columnwidth}{!}{
\begin{tabular}{ccccccccc}
    \toprule
         & NLI & SC & WSC & CR & SENT & SUM & MUL & AVG \\
    \midrule
        mT-Ko & 47.6 & 50.6 & 50.8 & 47.6 & 66.0 & \textbf{13.7} & 60.1 & 48.1\\
        mT-Ko-Trans & 51.4 & 54 & 52.3 & 48.9 & 74.1 & 13.3 & 65.8 & 51.4 \\
        mT-Ko-CI & \textbf{52.2} & \textbf{55.3} & \textbf{53.3} & \textbf{56.8} & \textbf{77.0} & \textbf{13.7} & \textbf{70.5} & \textbf{54.1} \\
    \bottomrule
\end{tabular}}
\caption{Performance of held-out P3 datasets with mT-Ko-Trans model \emph{(linguistic alignment only)}, and mT-Ko-CI model \emph{(linguistic and instructional format alignment)}. The best comparable performances are \textbf{bolded}. Details are on Appendix \ref{sec:instruction_alignment_breakdown}}
\label{table:CI-Trans}
\vspace{-5mm}
\end{table} 

\begin{figure}[t]
\centering
    \includegraphics[width=1\columnwidth]{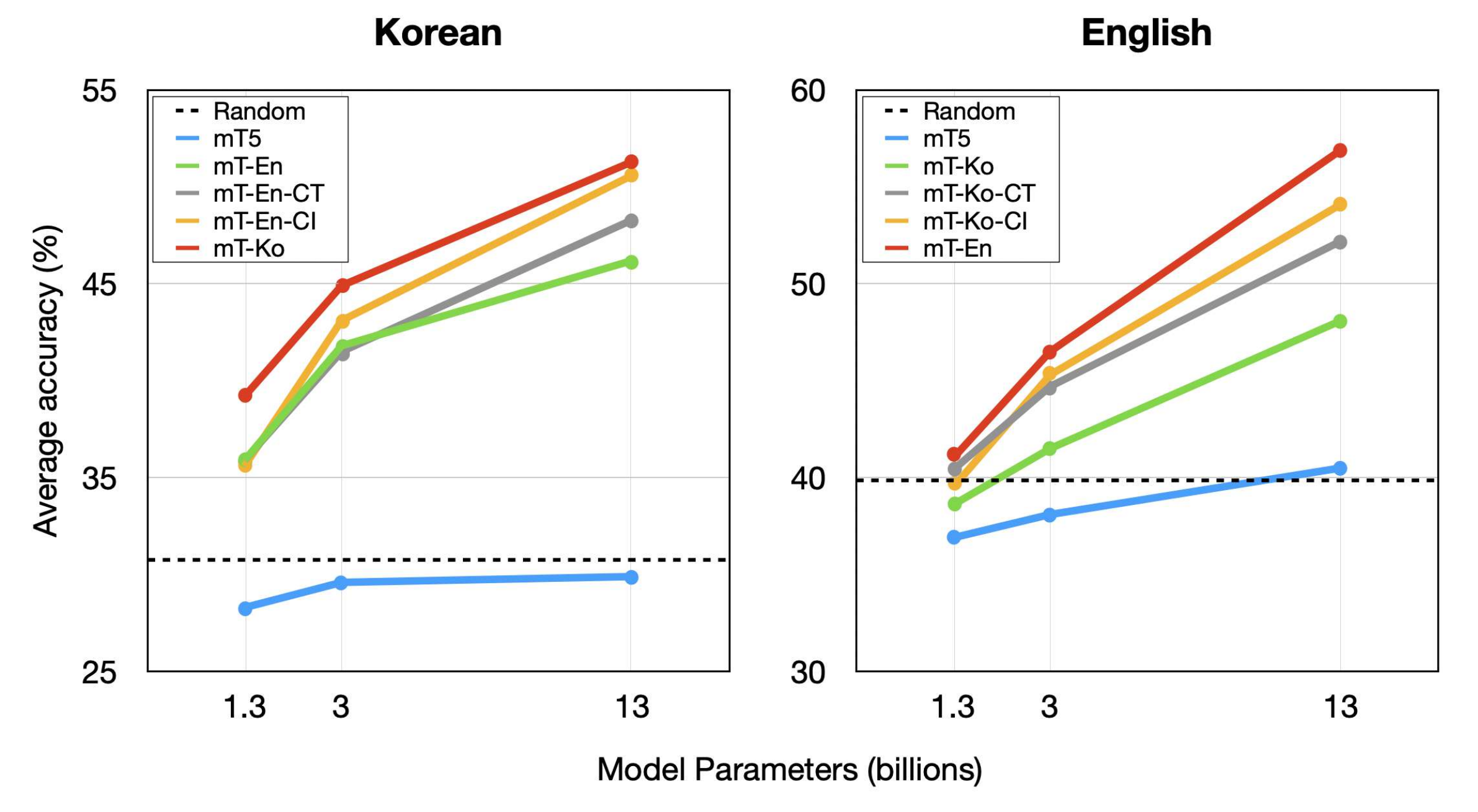}
\caption{Model performance vs. size. The random line represents the average score random choice in the options list for classification tasks, and the ROUGE-L score of a copy of input for generation tasks. Appendix \ref{sec:scale_up_breakdown} breaks down the performance by datasets.}
\label{fig:scaling_law}
\vspace{-5mm}
\end{figure}  

\subsection{Scaling Laws}
\label{sec:scaling law}
In our final ablation experiment, we investigate how the cross-lingual generalization of instruction tuning evolves with model size. Using the same model variants and cluster split as in Section \ref{sec:result}, we assess cross-lingual generalization performance across model sizes of 1.3B, 3B, and 13B.

Figure \ref{fig:scaling_law} illustrates the average performance of unseen tasks in both Korean and English. As model size increases, we observe performance improvements for all instruction-tuned models. Additionally, across various model sizes, models that incorporate cross-lingual templates exhibit higher performance, similar to the trends in the results of Section \ref{sec:result}. Particularly, for Korean evaluation, when the model size reaches 13B, mT-En-CI achieves comparable average performance to mT-Ko. In contrast, for English tasks, as the model size increases, the performance of mT-Ko-CI improves, but mT-En still exhibits performance differences. We conjecture that this trend may be attributed to the predominance of NLU tasks within the held-out task set. Given the profusion of NLU tasks within P3 compared to KORANI, mT-En, trained on P3, may possess an advantage in comprehending these held-out tasks in Korean and English. Detailed analyses of task-specific performance changes concerning model size are provided in Appendix \ref{sec:scale_up_breakdown}.

\section{Conclusion} 
Our research contributes to a deep understanding of the cross-lingual zero-shot generalization effect and its benefits by leveraging the novel KORANI meta-dataset to compare cross-lingual and monolingual instruction tuning directly. The experimental results indicate that cross-lingual instruction tuning can match or even exceed the performance of monolingual tuning. Our findings highlight the importance of training relevant data across diverse languages rather than strictly maintaining linguistic consistency in unseen tasks. The successful application of cross-lingual templates, which ensure consistency in both language and format, further validates the potential of cross-lingual instruction tuning. These discoveries present cross-lingual instruction tuning not just as an auxiliary strategy but as a potential alternative to monolingual methods, especially in low-resource language scenarios.

\section*{Limitations}
In this work, we primarily examine cross-lingual instruction tuning between Korean and English, which, while informative, provides a partial view of cross-lingual generalization due to the exclusion of other languages. Moreover, we utilized the mT5 model, a multilingual model trained in various languages, but more focused on English than Korean. Lastly, a difference in task cluster distribution between KORANI and P3 makes the result vague since the composition of datasets holds a significant effect. 


\section*{Acknowledgements}
We would like to express our heartfelt gratitude to Professor Moontae Lee for his invaluable guidance, discussions, and feedback throughout the course of this research. We also extend our sincere thanks to Joel Jang and Seonghyeon Ye for their thoughtful feedback on the paper.

\bibliography{anthology,custom}
\bibliographystyle{acl_natbib}

\appendix

\label{sec:appendix}

\section{Dataset}
\subsection{Public Release} 
To foster active research on instruction tuning in the Korean community, we have made CSV files available on GitHub, containing instruction-templated inputs and outputs for benchmarks that have free copyright of derivative works. For the five out of 51 datasets that have limited copyright of derivative works, we have provided a method to download the data and preprocess code with underlying templates. Our objective is to enable researchers to readily utilize these datasets for instruction tuning research while respecting copyright laws\footnote{For detailed information about data sources and license information, please refer to the following link: \url{https://github.com/CHLee0801/KORANI-Instruction-Tuning}.}.

\subsection{Example of Raw Data Transformations}
\label{sec:raw data transformation}
Open-source datasets included in KORANI are typically characterized by predefined tasks and various labels. We utilize these labels, or generate new ones, to create templates. Even if the original purpose of the dataset differs, we leverage the labels to develop turn-around tasks, similar to approaches used in \citet{T0} and \citet{Flan}. For instance, in the case of dialog tasks, such as AIHub TOD, the presence of labels indicating the topic of conversation enabled us to create topic classification tasks. For summarization tasks, we create labels by extracting keywords using KeyBERT and classify topics of the document using the TF-IDF algorithm. We then utilize these labels to enrich the instructions, thereby incorporating contextual information about the conversation's topic and enhancing the overall comprehensiveness of the instructions.

\subsection{P3 Datasets and Task Taxonomy}
\label{sec:p3_figure}
\begin{figure*}[ht!]
\centering
    \includegraphics[width=0.8\textwidth]{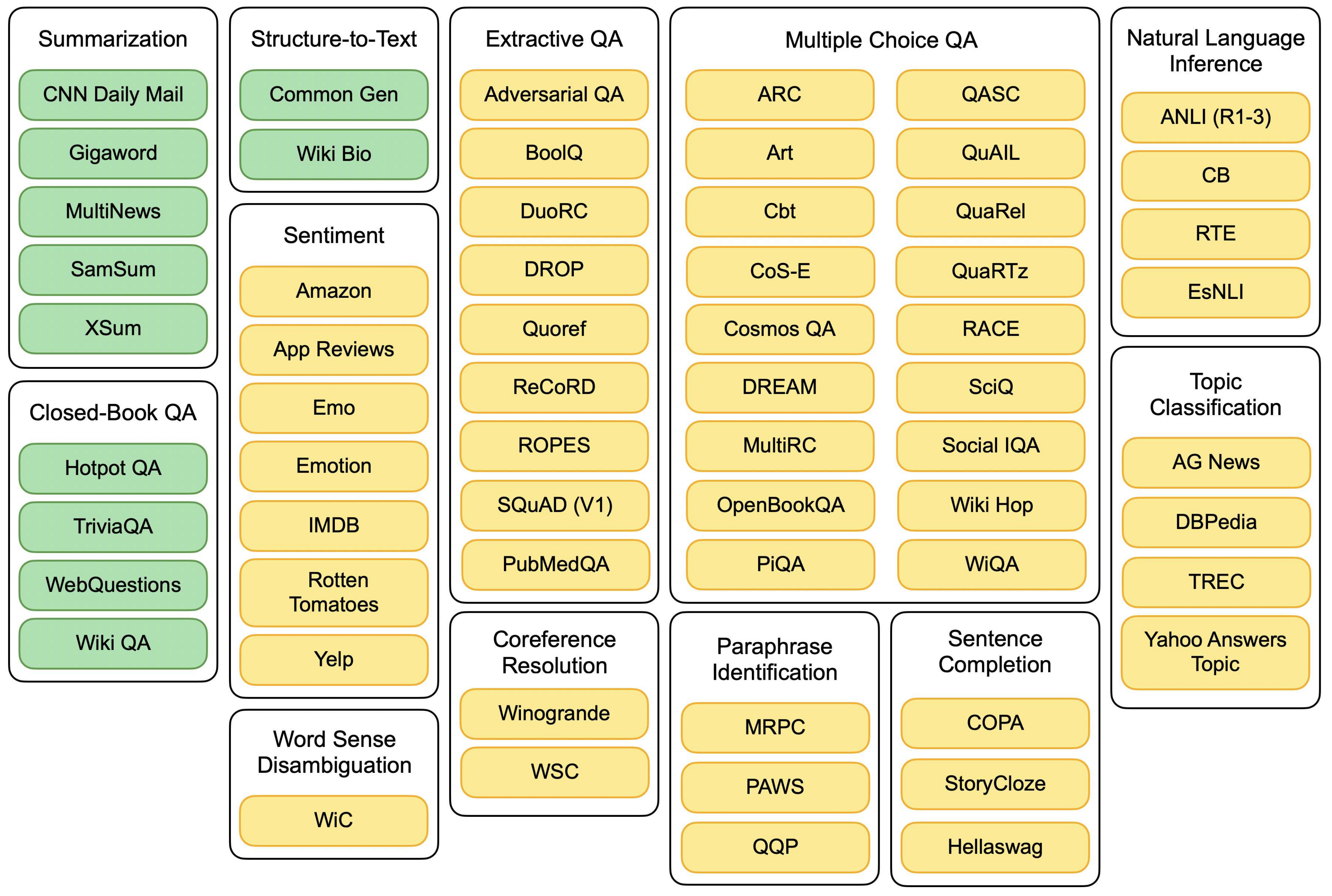}
\caption{\textsc{P3} datasets and task taxonomy. Green datasets are NLG datasets, Yellow datasets are NLU datasets. We follow task categorization from \citet{T0}}
\label{fig:P3 Datasets}
\end{figure*}

Please refer to Figure \ref{fig:P3 Datasets}.

\subsection{Cross-lingual Templates Generation}
\label{sec:cross-lingual templates generation}
To encompass linguistic attributes for cross-lingual templates, instead of just translating corresponding task templates from another language, a concerted effort is exerted to extract the underlying semantics and salient components inherent to the templates. Subsequently, these components are meticulously organized and structured in a manner that not only accentuates the intrinsic qualities of the target language but also ensures the retention of task-specific characteristics. Within the context of generating cross-lingual templates, a fundamental aspect involves the meticulous alignment of structural elements across languages. Furthermore, the configuration of choices for classification is adapted to the auxiliary language, unless the task inherently demands distinct sentence or phrase-based choices for each instance—illustratively observed in tasks like sentence completion and multiple-choice QA. The strategic deployment of demonstrative pronouns and the method of incorporating meta-data are thoughtfully tailored by the structural framework of the templates in question, thereby underlining the significance of syntactic considerations. Please refer to Section \label{sec:examples_of_ci_trans} for specific examples.

\subsection{Illustrative examples of cross-lingual templates and translated templates}
\label{sec:examples_of_ci_trans}

The examples below demonstrate the original English templates from Promptsource, cross-lingual templates for mT-Ko-Trans, and mT-Ko-CI. mT-Ko-Trans is a translated version of English templates, and the structural format is identical as well, while mT-Ko-CI is more aligned with templates in KORANI datasets. Table \ref{table:xsum-cross-template}, Table \ref{table:wsc-cross-template}, and Table \ref{table:emotion-cross-template} illustrate the examples of cross-lingual templates from P3 datasets. Moreover, Table \ref{table:document-qa-cross-template}, Table \ref{table:klue-nli-cross-template}, and Table \ref{table:kobest-wic-cross-template} illustrate the examples of cross-lingual templates from KORANI datasets. All meta-data of the dataset are represented in double brackets.

\begin{table}[]
\centering
    \resizebox{\columnwidth}{!}{
\begin{tabular}{ll}
    \toprule
       & P3 Templates\\
    \midrule
        \textbf{Input} & \{\{document\}\}\textbackslash n\textbackslash n ===\textbackslash n\textbackslash nWrite a summary of the text above : \\
        \textbf{Output} & \{\{summary\}\}\\
    \bottomrule
        & Translated Templates\\
    \midrule
        \textbf{Input} & \{\{document\}\}\textbackslash n\textbackslash n ===\textbackslash n\textbackslash n위 글을 영어로 요약하시오. \\
        \textbf{Output} & \{\{summary\}\}\\
    \bottomrule
        & Cross-lingual Templates\\
    \midrule
        \textbf{Input} & 다음은 글을 읽고 요약하는 문제입니다.\textbackslash n \\
        & \{\{document\}\}\textbackslash n위 글을 영어로 요약하세요. \\ 
        \textbf{Output} & \{\{summary\}\}\\
    \bottomrule
\end{tabular}}
\caption{Instruction examples of XSum : Summarization.}
\label{table:xsum-cross-template}
\end{table} 

\begin{table}[]
\centering
    \resizebox{\columnwidth}{!}{
\begin{tabular}{ll}
    \toprule
       & P3 Templates\\
    \midrule
        \textbf{Input} & \{\{text\}\} In the previous sentence, does the pronoun \\
        & "\{\{span2\_text\}\}" refer to \{\{span1\_text\}\}? Yes or no? \\
        \textbf{Choices} & Yes ||| no \\
        \textbf{Output} & \{\{ answer\_choices [label] \}\} \\
    \bottomrule
        & Translated Templates\\
    \midrule
        \textbf{Input} & \{\{text\}\} 이전 문장에서 단어 "\{\{span1\_text\}\}" \\
        & 는 "\{\{span2\_text\}\}"를 참조하는가? 예, 아니오 \\
        \textbf{Choices} & 예 ||| 아니오 \\
        \textbf{Output} & \{\{ answer\_choices [label] \}\} \\
    \bottomrule
        & Cross-lingual Templates\\
    \midrule
        \textbf{Input} & 글에서 같은 것을 의미하는 다른 두 단어는 서로를\\
        & 참조하는 관계이다. 문장: \{\{text\}\}\textbackslash n위 문장에서 단어\\
        & \{\{span1\_text\}\}와 "\{\{span2\_text\}\}" 의 뜻이 같은가? \\ 
        \textbf{Choices} & 예 ||| 아니오 \\
        \textbf{Output} & \{\{ answer\_choices [label] \}\} \\
    \bottomrule
\end{tabular}}
\caption{Instructions examples of WSC : Coreference Resolution.}
\label{table:wsc-cross-template}
\end{table} 

\begin{table}[]
\centering
    \resizebox{\columnwidth}{!}{
\begin{tabular}{ll}
    \toprule
       & P3 Templates \\
    \midrule
        \textbf{Input} & \{\{text\}\}. What is the emotion expressed in this message? \\
        \textbf{Choices} & sadness ||| joy ||| love ||| anger ||| fear ||| surprise \\
        \textbf{Output} & \{\{ answer\_choices [label] \}\} \\
    \bottomrule
        & Translated Templates \\
    \midrule
        \textbf{Input} & \{\{text\}\} 이 메세지에 나타난 감정은 무엇인가? \\
        \textbf{Choices} & 슬픔 ||| 기쁨 ||| 사랑 ||| 화남 ||| 공포 ||| 놀람 \\
        \textbf{Output} & \{\{ answer\_choices [label] \}\} \\
    \bottomrule
        & Cross-lingual Templates\\
    \midrule
        \textbf{Input} & 감정 분류 태스크이다.\textbackslash n\{\{text\}\}\textbackslash n \\
        & 위에서 확인할 수 있는 사람의 감정을 알려줘. \\ 
        \textbf{Choices} & 슬픔 ||| 기쁨 ||| 사랑 ||| 화남 ||| 공포 ||| 놀람 \\
        \textbf{Output} & \{\{ answer\_choices [label] \}\} \\
    \bottomrule
\end{tabular}}
\caption{Instruction examples of Emotion : Sentiment.}
\label{table:emotion-cross-template}
\end{table} 

\begin{table}[]
\centering
    \resizebox{\columnwidth}{!}{
\begin{tabular}{ll}
    \toprule
       & KORANI Templates \\
    \midrule
        \textbf{Input} & 다음은 보기가 주어져서 정답을 고르는 기계독해 \\
        & 문제이다. 답을 \{\{choices\}\} 중에서 고르시오.\textbackslash n \\
        & \{\{context\}\}\textbackslash n\{\{question\}\} \\
        \textbf{Choices} & choices[0] ||| choices[1] ||| choices[2] ||| choices[3] \\
        \textbf{Output} & \{\{ answer\_choices [label] \}\} \\
    \bottomrule
        & Cross-lingual Templates\\
    \midrule
        \textbf{Input} & Read the following context and choose the best option to \\
        & answer the question. \textbackslash nContext: \{\{context\}\}\textbackslash nQuestion:\\
        & \{\{question\}\}\textbackslash nOptions:\textbackslash n \{\{choices\}\} \\
        \textbf{Choices} & choices[0] ||| choices[1] ||| choices[2] ||| choices[3] \\
        \textbf{Output} & \{\{ answer\_choices [label] \}\} \\
    \bottomrule
\end{tabular}}
\caption{Instruction examples of Document QA : Multiple Choice QA.}
\label{table:document-qa-cross-template}
\end{table} 

\begin{table}[]
\centering
    \resizebox{\columnwidth}{!}{
\begin{tabular}{ll}
    \toprule
       & KORANI Templates \\
    \midrule
        \textbf{Input} & 자연어 추론 문제이다. 이 문제는 전제가 참이라고 \\
        & 가정할 때, 가설의 내용이 참(함의)인지, 거짓(모순)인지,\\
        & 혹은 알 수 없는지(무관)에 따라 관계가 분류된다. 전제와 \\
        & 가설의 관계를 유추하라.\textbackslash n전제: \{\{premise\}\}\textbackslash n \\
        & 가설: \{\{hypothesis\}\}\textbackslash n선택지: \{\{choices\}\} \\
        \textbf{Choices} & 함의 ||| 모순 ||| 무관 \\
        \textbf{Output} & \{\{ answer\_choices [label] \}\} \\
    \bottomrule
        & Cross-lingual Templates\\
    \midrule
        \textbf{Input} & \{\{premise\}\} Using only the above \\
        & description and what you know about the \\
        & world, ""\{\{hypothesis\}\}"" is definitely \\
        & correct, incorrect, or inconclusive? \\
        \textbf{Choices} & Correct ||| Inconclusive ||| Incorrect \\
        \textbf{Output} & \{\{ answer\_choices [label] \}\} \\
    \bottomrule
\end{tabular}}
\caption{Instruction examples of KLUE NLI : Natural Language Inference.}
\label{table:klue-nli-cross-template}
\end{table} 

\begin{table}[]
\centering
    \resizebox{\columnwidth}{!}{
\begin{tabular}{ll}
    \toprule
       & KORANI Templates \\
    \midrule
        \textbf{Input} & 아래 두 문장에서 [\{\{word\}\}]의 \\
        & 뜻이 같은지 판별하시오.\textbackslash n\{\{sentence1\}\}\textbackslash n\{\{sentence2\}\} \\
        & \textbackslash n선택지: \{\{choices\}\} \\
        \textbf{Choices} & 예 ||| 아니오 \\
        \textbf{Output} & \{\{ answer\_choices [label] \}\} \\
    \bottomrule
        & Cross-lingual Templates\\
    \midrule
        \textbf{Input} & Does the word [\{\{word\}\}] have \\
        & the same meaning in these two sentences?\textbackslash n \\
        & \{\{sentence1\}\}\textbackslash n\{\{sentence2\}\}\textbackslash n\{\{choices\}\} \\
        \textbf{Choices} & Yes ||| No \\
        \textbf{Output} & \{\{ answer\_choices [label] \}\} \\
    \bottomrule
\end{tabular}}
\caption{Instruction examples of Kobest WiC : Word Sense Disambiguation.}
\label{table:kobest-wic-cross-template}
\end{table}

\section{Training and evalauation}
\subsection{training}
\label{sec:training_detail}
The second group consists of three tasks: sentiment analysis, summarization, and multiple-choice QA. We choose sentiment analysis as a held-out task because it has the potential to discern whether the model effectively comprehends semantic nuances across languages. We also hold out summarization to verify the extent of task generalization within generative tasks. Lastly, the decision to hold out multiple-choice QA stems from the intricate nature of the task's choices, which demand a nuanced understanding of linguistic subtleties beyond mere structural or template patterns. 

We truncate input and target sequences to 768 and 256 tokens, respectively. We train all models with a batch size of 64 using AdamW Optimizer with a learning rate of 1e-5. We also train all models for 1 epoch and save checkpoints for every 600 steps to select checkpoints for evaluation.

For validation, we sample 100 examples from the validation splits of each training dataset. We measure the performance of each dataset and aggregate them to perform checkpoint selection. This approach avails our experiment in a true zero-shot setting, as we do not use any examples from held-out tasks for checkpoint selection.

\section{More Analysis on Cross-Lingual Instruction Tuning}
\label{sec:cross_detail}

We claim the performance enhancement stems from the inclusion of relevant tasks in the training phase, with the enhancement increasing when the language is aligned. For example, we hypothesize that including the "piqa" dataset in mT-En training boosts performance on sentence completion tasks, as "piqa", a multiple-choice QA dataset, closely resembles sentence completion tasks. Furthermore, for sentiment analysis, mT-Ko-CT and mT-Ko-CI demonstrate superior performance in English evaluations, likely due to the inclusion of the hatespeech task from the KORANI datasets. These datasets, which are designed to identify sentence toxicity, may align well with sentiment analysis tasks.

\begin{table}[t]
\centering
    \resizebox{0.9\columnwidth}{!}{
\begin{tabular}{l|cccc}
\hline
\textbf{Models} & \textbf{NLI} & \textbf{SC} & \textbf{WSD} & \textbf{CR} \\
\hline
\hline
mT5 & 33.5 & 40.2 & 49.3 & 1.2 \\
mT-Ko & 42.5 & 53.1 & 56.1 & 23.3 \\
mT-Bi & 44.5 & 55.8 & 55.0 & \underline{32.2} \\
mT-Bi-CT & 46.8 & \textbf{58.2} & 58.0 & 31.1 \\
mT-Bi+ & \underline{58.4} & \underline{57.7} & \underline{59.6} & 31.9 \\
mT-Bi+-CT & \textbf{76.7} & 56.7 & \textbf{61.8} & \textbf{32.4} \\
\hline
\end{tabular}
}
\caption{Performance of Bilingual Instruction Tuning on KORANI Evaluation Benchmarks (Natural Language Inference, Sentence Completion, Word Sense Disambiguation, Coreference Resolution) for Different Models.}
\label{table:bi_add}
\end{table}

\section{Additional Experiment on Bilingual Instruction Tuning}
\label{sec:add-bi}
We conducted an extra experiment on mT-Bi (CT) for specific tasks on Table \ref{table:bi_add}. The trend observed aligns with the findings presented in Figure \ref{fig:bi-instruction-tuning}.

\section{Results Breakdown}
This section shows the full results for all datasets we evaluate. We show the performance of the models using randomly chosen three templates per dataset with the best performance on the dev set. All results are average scores of three templates. For efficient evaluation, we sample a maximum of 1000 instances for all generative datasets. These include summarization datasets for KORANI and P3, and coreference resolution datasets for KORANI. We use greedy search for all generative tasks.
\subsection{Zero-shot cross-lingual generalization performance breakdown}
\label{sec:crosslingual_breakdown}
Table \ref{table:korani_crosslingual_generalization} and Table \ref{table:p3_crosslingual_generalization} break down the scores of zero-shot cross-lingual generalization performance of KORANI and P3 respectively. 

\begin{table*}[h!]
\centering
    \resizebox{0.9\textwidth}{!}{\begin{tabular}{cccccccccccccc}
    \toprule
    \multirow{3}{*}{\textbf{Model}} & \multicolumn{5}{c}{Sentiment} & \multirow{3}{*}{\textbf{Avg.}} & \multicolumn{5}{c}{Summarization} & \multirow{3}{*}{\textbf{Avg.}} & Mul. QA
    \\ \cmidrule(lr){2-6} \cmidrule(lr){8-12} \cmidrule(lr){14-14} 
    & \textbf{AIHub} & \textbf{Kobest} & \multirow{2}{*}{\textbf{NSMC}} & \textbf{Naver} & \textbf{Sosang} & & \multirow{2}{*}{\textbf{Book}}& \textbf{Dacon} & \textbf{Doc.} & \textbf{Doc.} & \multirow{2}{*}{\textbf{Report}} & & \textbf{Doc.} \\
    & \textbf{Emo} & \textbf{Sentiment} & & \textbf{Shopping} & \textbf{Sentiment} & & & \textbf{News} & \textbf{Edi.} & \textbf{News} & & & \textbf{QA} \\
    \midrule
    mT5 & 18.2 & 50.4 & 54.8 & 50 & 33.5 & 41.4 & 5.8 & 6.8 & 5.6 & 8 & 4.8 & 6.2 & 37.4 \\
    \midrule
    mT-En & 53.5 & 54.5 & 51.9 & 51 & 35.3 & 49.2 & 26.3 & 25.1 & 15.8 & 23.3 & \textbf{19.2} & \underline{21.9} & 84.7 \\
    mT-En-CT & \textbf{55} & 83.4 & 63.4 & 62.1 & 46.2 & 62 & 23.4 & 23.6 & 15.3 & 20.8 & 17.9 & 20.2 & \textbf{86.4} \\
    mT-En-CI & 46.7 & 90.7 & \textbf{81.6} & \textbf{87.2} & \textbf{55.8} & \underline{72.4} & 17.6 & 20.8 & 13.8 & 20.1 & 14.3 & 17.3 & 85.6 \\
    \midrule
    mT-Ko & 54.4 & \textbf{96.4} & 80.2 & 84.1 & 55.5 & \textbf{74.1} & \textbf{30} & \textbf{28.6} & \textbf{17.5} & \textbf{24.4} & 18.6 & 
    \textbf{23.8} & \underline{86.2} \\
    \bottomrule
    \end{tabular}}
    \resizebox{0.7\textwidth}{!}{
    \begin{tabular}{cccccccccc}
    \toprule
    \multirow{3}{*}{\textbf{Model}} & \multicolumn{2}{c}{NLI} & \multirow{3}{*}{\textbf{Avg.}} & \multicolumn{2}{c}{Sent. Comp.} & \multirow{3}{*}{\textbf{Avg.}} & Coref. Resol & WSD & \multirow{3}{*}{\textbf{Total Avg.}}
    \\ \cmidrule(lr){2-3} \cmidrule(lr){5-6} \cmidrule(lr){8-8} \cmidrule(lr){9-9} & \textbf{KLUE} & \multirow{2}{*}{\textbf{KorNLI}} & & \textbf{Kobest} & \textbf{Kobest} & & \multirow{2}{*}{\textbf{NIKL Coref}} & \textbf{Kobest} &  \\
    & \textbf{NLI}& & & \textbf{Copa} & \textbf{Hellas.} & & & \textbf{WiC} & \\
    \midrule
    mT5 & 33.4 & 33.6 & 33.5 & 53.3 & 27 & 40.2 & 1.2 & 49.3 & 29.9 \\
    \midrule
    mT-En & 39.7 & 35.3 & 37.5 & \underline{66.9} & 39.7 & 53.3 & \textbf{23.5} & \underline{53.2} & 46.2 \\
    mT-En-CT & 39.4 & 34.9 & 37.2 & 66.6 & \underline{43.2} & \underline{54.9} & 21 & \textbf{56.1}  & 48.3 \\
    mT-En-CI & \textbf{58.7} & \textbf{52.1} & \textbf{55.4} & \textbf{80.8} & 34  & \textbf{57.4} & 22 & 52.7 & \textbf{51.8} \\
    \midrule
    mT-Ko & \underline{44.7} & \underline{40.3} & \underline{42.5} & 62.1 & \textbf{44} & 53.1 & \underline{23.3} & \textbf{56.1}  & \underline{51.3} \\
    \bottomrule
    \end{tabular}}
\caption{KORANI zero-shot cross-lingual generalization performance breakdown. The best comparable performances are \textbf{bolded} and second best \underline{underlined}. }
\label{table:korani_crosslingual_generalization}
\end{table*}

\begin{table*}[h!]
    \resizebox{\textwidth}{!}{\begin{tabular}{cccccccccccccccccc}
    \toprule
    \multirow{2}{*}{\textbf{Model}}  & \multicolumn{4}{c}{Sentiment} & \multirow{2}{*}{\textbf{Avg.}} & \multicolumn{5}{c}{Summarization} & \multirow{2}{*}{\textbf{Avg.}} & \multicolumn{5}{c}{Multiple-Choice QA} & \multirow{2}{*}{\textbf{Avg.}}
    \\ \cmidrule(lr){2-5} \cmidrule(lr){7-11} \cmidrule(lr){13-17} & \textbf{Emot.} & \textbf{Rot. Tom.} & \textbf{Ama.} & \textbf{IMDB} & & \textbf{Mul.News} & \textbf{Sam.} & \textbf{CNN.} & \textbf{XSum} & \textbf{Giga.} & & \textbf{Dream} & \textbf{Mul.RC} & \textbf{PiQA} & \textbf{QASC} & \textbf{RACE} & \\
    \midrule
    mT5 & 33.4 & 50.4 & 49.7 & 53.5 & 46.8 & 6.4  & 3.1 & 3.8 & 2.4 & 2.4 & 3.6 & 36.6 & 57.2 & 52.4 & 49.2 & 27.3 & 44.5\\
    \midrule
    mT-Ko & 35.8 & 67.7 & \underline{84.9} & 75.7 & 66 & \textbf{7.8} & 11 & \textbf{15.1} & \underline{11.9} & \textbf{22.9} & \underline{13.7} & 59.1 & 65.3 & 57.2 & 74.6 & 44.2 & 60.1 \\
    mT-Ko-CT & \underline{43.6} & \underline{84.1} & 76.4 & \underline{77.7} & 70.5 & 6.9 & 9 & \underline{14.5} & 11.8 & 20.2 & 12.5 & \underline{74.8} & 77.8 & \textbf{62.3} & \underline{91.4} & 46 & \underline{70.5} \\
    mT-Ko-CI & \textbf{49.4} & 77.8 & \textbf{91.2} & \textbf{89.6} & \textbf{77} & 6.7 & \textbf{20.2} & 13.1 & 10.3 & 18.3 & \underline{13.7} & 69 & \textbf{79.8} & 58.2 & 91 & \textbf{54.6} & \underline{70.5} \\
    \midrule
    mT-En & 40.2 & \textbf{84.5} & 80 & \underline{77.7} & \underline{70.6} & \underline{7} & \underline{17.1} & 14.3 & \textbf{12.7} & \underline{22.4} & \textbf{14.7} & \textbf{77.6} & \underline{78.2} & \underline{61.1} & \textbf{93.8} & \underline{47.7} & \textbf{71.7} \\
    \bottomrule
    \end{tabular}}
    \resizebox{\textwidth}{!}{\begin{tabular}{ccccccccccccccccc}
    \toprule
    \multirow{2}{*}{\textbf{Model}} & \multicolumn{6}{c}{Natural Language Inference} & \multirow{2}{*}{\textbf{Avg.}} & \multicolumn{3}{c}{Sentence Completion} & \multirow{2}{*}{\textbf{Avg.}} & \multicolumn{2}{c}{Coref. Resol.} & \multirow{2}{*}{\textbf{Avg.}} & WSD & \multirow{2}{*}{\textbf{Total Avg.}}
    \\ \cmidrule(lr){2-7} \cmidrule(lr){9-11} \cmidrule(lr){13-14} \cmidrule(lr){16-16} & \textbf{RTE} & \textbf{CB} & \textbf{AN. R1} & \textbf{AN. R2} & \textbf{AN. R3} & \textbf{EsNLI} & & \textbf{COPA} & \textbf{Hellasw.} & \textbf{StoryC.} & & \textbf{Winogr.} & \textbf{WSC} & & \textbf{WiC} &  \\
    \midrule
    mT5 & 47.3 & 50.6 & 32.8 & 33.4 & 33 & 33.9 & 38.5 & 56 & 26 & 49.9 & 44 & 49.2 & \underline{63.5} & 56.4 & 50 & 40.5 \\
    \midrule
    mT-Ko & 65.3 & 75 & 36.2 & 34 & \underline{37.5} & \underline{37.5} & 47.6 & 63.7 & 28.9 & 59.1 & 50.6 & 51.9 & 43.3 & 47.6 & 50.8 & 48.1 \\
    mT-Ko-CT & 76.8 & \underline{83.3} & \underline{38.2} & \underline{34.6} & 37.4 & 34.3 & 50.8 & \underline{66.3} & 31.8 & \underline{67.6} & 55.2 & \underline{52.5} & 58 & 55.3 & 50.8 & 52.2 \\
    mT-Ko-CI & \underline{79.5} & 79.2 & 33.6 & 33.8 & 33.0 & \textbf{53.3} & \underline{52.2} & 63.3 & \textbf{35.4} & 67.3 & \underline{55.3} & 51.6 &  61.9 & \underline{56.8} & \textbf{53.3} & \underline{54.1} \\
    \midrule
    mT-En & \textbf{81.3} & \textbf{85.1} & \textbf{40.5} & \textbf{36.5} & \textbf{40.2} & 34.4 & \textbf{53} & \textbf{85.3} & \underline{34.9} & \textbf{94.7} & \textbf{71.6} & \textbf{61.6} & \textbf{67} & \textbf{64.3} & \underline{52.6} & \textbf{56.9} \\
    \bottomrule
    \end{tabular}}
\caption{P3 zero-shot cross-lingual generalization performance breakdown. The best comparable performances are \textbf{bolded} and second best \underline{underlined}. }
\label{table:p3_crosslingual_generalization}
\end{table*}

\subsection{Bilingual Instruction Tuning Performance Breakdown}
\label{sec:bilingual_breakdown}
Table \ref{table:korani_bilingual_generalization} and Table \ref{table:p3_bilingual_generalization} break down the scores of bilingual instruction tuning performance of KORANI and P3 respectively. 

\begin{table*}[h!]
\centering
\resizebox{0.9\textwidth}{!}{\begin{tabular}{cccccccccccccc}
    \toprule
    \multirow{3}{*}{\textbf{Model}} & \multicolumn{6}{c}{Sentiment} & \multicolumn{6}{c}{Summarization} & Mul. QA
    \\ \cmidrule(lr){2-7} \cmidrule(lr){8-13} \cmidrule(lr){14-14} 
    & \textbf{AIHub} & \textbf{Kobest} & \multirow{2}{*}{\textbf{NSMC}} & \textbf{Naver} & \textbf{Sosang} & \multirow{2}{*}{\textbf{Avg.}} & \multirow{2}{*}{\textbf{Book}}& \textbf{Dacon} & \textbf{Doc.} & \textbf{Doc.} & \multirow{2}{*}{\textbf{Report}} & \multirow{2}{*}{\textbf{Avg.}} & \textbf{Doc.} \\
    & \textbf{Emo} & \textbf{Sentiment} & & \textbf{Shopping} & \textbf{Sentiment} & & & \textbf{News} & \textbf{Edi.} & \textbf{News} & & & \textbf{QA} \\
    \midrule
    mT5 & 18.2 & 50.4 & 54.8 & 50 & 33.5 & 41.4 & 5.8 & 6.8 & 5.6 & 8 & 4.8 & 6.2 & 37.4 \\
    \midrule
    mT-Ko & \textbf{54.4} & \textbf{96.4} & 80.2 & \underline{84.1} & \underline{55.5} & \underline{74.1} & 30 & 28.6 & 17.5 & 24.4 & 18.6 & 23.8 & 86.2 \\
    mT-Bi & \underline{52.5} & \underline{96.2} & 76.8 & 72.8 & 51.7 & 70 & 30.9 & 29.5 & \underline{18.7} & 23.9 & 19.1 & 24.4 & 87.4 \\
    mT-Bi+ & 54 & 96.1 & \underline{80.7} & 78.9 & 54.9 & 72.9 & \textbf{32.6} & \textbf{36.5} & \textbf{21.2} & \textbf{32.2} & \textbf{24.5} & \textbf{29.4} & \underline{88.9} \\
    mT-Bi+-CT & 49.6 & 96 & \textbf{84.5} & \textbf{92.2} & \textbf{61.2} & \textbf{76.7} & \underline{32.4} & \underline{32.9} & \textbf{21.2} & \underline{27.9} & \underline{23.4} & \underline{27.6} & \textbf{89.5} \\
    \bottomrule
    \end{tabular}}
\resizebox{0.7\textwidth}{!}{
    \begin{tabular}{cccccccccc}
    \toprule
    \multirow{3}{*}{\textbf{Model}} & \multicolumn{3}{c}{NLI} & \multicolumn{3}{c}{Sent. Comp.} & Coref. Resol & WSD & \multirow{3}{*}{\textbf{Total Avg.}}
    \\ \cmidrule(lr){2-4} \cmidrule(lr){5-7} \cmidrule(lr){8-8} \cmidrule(lr){9-9} & \textbf{KLUE} & \multirow{2}{*}{\textbf{KorNLI}} & \multirow{2}{*}{\textbf{Avg.}} & \textbf{Kobest} & \textbf{Kobest} & \multirow{2}{*}{\textbf{Avg.}} & \multirow{2}{*}{\textbf{NIKL Coref}} & \textbf{Kobest} &  \\
    & \textbf{NLI}& & & \textbf{Copa} & \textbf{Hellas.} & & & \textbf{WiC} & \\
    \midrule
    mT5 & 33.4 & 33.6 & 33.5 & 53.3 & 27 & 40.2 & 1.2 & 49.3 & 29.9 \\
    \midrule
    mT-Ko & 44.7 & 40.3 & 42.5 & 62.1 & 44 & 53.1 & 23.3 & 56.1 & 51.3 \\
    mT-Bi & 46.6 & 42.4 & 44.5 & \underline{66} & 45.5 & 55.8 & \underline{32.2} & 55 & 52.6 \\
    mT-Bi+ & \underline{55} & \underline{61.8} & \underline{58.4} & \textbf{67.7} & \textbf{47.6} & \textbf{57.7} & 31.9 & \underline{59.6} &  \underline{56.5} \\
    mT-Bi+-CT & \textbf{82.7} & \textbf{70.6} & \textbf{76.7} & \textbf{67.7} & \underline{45.6} & \underline{56.7} & \textbf{32.4} & \textbf{61.8} & \textbf{59.7} \\
    \bottomrule
    \end{tabular}}
\caption{KORANI bilingual instruction tuning performance breakdown. The best comparable performances are \textbf{bolded} and second best \underline{underlined}. }
\label{table:korani_bilingual_generalization}
\end{table*}

\begin{table*}[h!]
\resizebox{\textwidth}{!}{\begin{tabular}{cccccccccccccccccc}
    \toprule
    \multirow{2}{*}{\textbf{Model}}  & \multicolumn{5}{c}{Sentiment} & \multicolumn{6}{c}{Summarization} & \multicolumn{6}{c}{Multiple-Choice QA}
    \\ \cmidrule(lr){2-6} \cmidrule(lr){7-12} \cmidrule(lr){13-18} & \textbf{Emot.} & \textbf{Rot. Tom.} & \textbf{Ama.} & \textbf{IMDB} & \textbf{Avg.} & \textbf{Mul.News} & \textbf{Sam.} & \textbf{CNN.} & \textbf{XSum} & \textbf{Giga.} & \textbf{Avg.} & \textbf{Dream} & \textbf{Mul.RC} & \textbf{PiQA} & \textbf{QASC} & \textbf{RACE} & \textbf{Avg.}\\
    \midrule
    mT5 & 33.4 & 50.4 & 49.7 & 53.5 & 46.8 & 6.4  & 3.1 & 3.8 & 2.4 & 2.4 & 3.6 & 36.6 & 57.2 & 52.4 & 49.2 & 27.3 & 44.5\\
    \midrule
    mT-En & 40.2 & 84.5 & 80 & 77.7 & 70.6 & 7 & 17.1 & 14.3 & 12.7 & 22.4 & 14.7 & 77.6 & 78.2 & 61.1 & 93.8 & \underline{47.7} & 71.7 \\
    mT-Bi & 47.2 & 83.5 & 80 & \underline{77.8} & 72.1 & 7.6 & 15.8 & 13.8 & 13.1 & \textbf{28.4} & \underline{15.7} & 80.1 & \textbf{81.2} & 60.7 & 94.7 & 47.2 & 72.8 \\
    mT-Bi+ & \textbf{54.3} & \underline{87.6} & \underline{81} & \textbf{80.3} & \underline{75.8} & \textbf{10.9} & \textbf{25} & \underline{18.7} & \underline{14.1} & \underline{27.9} & 13.7 & \underline{80.9} & 80.6 & \underline{61.9} & \textbf{95.4} & \textbf{48.1} & \underline{73.4} \\
    mT-Bi+-CT & \underline{51.8} & \textbf{89.9} & \textbf{96.7} & \textbf{80.3} & \textbf{79.7} & \underline{10.7} & \underline{21.5} & \textbf{19} & \textbf{14.4} & \underline{27.9} & \textbf{18.7} & \textbf{81.7} & \underline{80.8} & \textbf{62.2} & \underline{95.2} & 47.5 & \textbf{73.5} \\
    \bottomrule
    \end{tabular}}
    \resizebox{\textwidth}{!}{\begin{tabular}{ccccccccccccccccc}
    \toprule
    \multirow{2}{*}{\textbf{Model}} & \multicolumn{7}{c}{Natural Language Inference} & \multicolumn{4}{c}{Sentence Completion} & \multicolumn{3}{c}{Coref. Resol.} & WSD & \multirow{2}{*}{\textbf{Total Avg.}}
    \\ \cmidrule(lr){2-8} \cmidrule(lr){9-12} \cmidrule(lr){13-15} \cmidrule(lr){16-16} & \textbf{RTE} & \textbf{CB} & \textbf{AN. R1} & \textbf{AN. R2} & \textbf{AN. R3} & \textbf{EsNLI} & \textbf{Avg.} & \textbf{COPA} & \textbf{Hellasw.} & \textbf{StoryC.} & \textbf{Avg.} & \textbf{Winogr.} & \textbf{WSC} & \textbf{Avg.} & \textbf{WiC} &  \\
    \midrule
    mT5 & 47.3 & 50.6 & 32.8 & 33.4 & 33 & 33.9 & 38.5 & 56 & 26 & 49.9 & 44 & 49.2 & 63.5 & 56.4 & 50 & 40.5 \\
    \midrule
    mT-En & 81.3 & 85.1 & 40.5 & 36.4 & 40.2 & 34.4 & 53 & 85.3 & 34.9 & \underline{94.7} & 71.6 & 61.6 & 67 & 64.3 & \underline{52.6} & 56.9 \\
    mT-Bi & 83.9 & \textbf{88.1} & \underline{43.2} & \underline{36.8} & \underline{41.6} & 43.6 & 56.2 & \underline{85.7} & \underline{36} & 94 & \underline{71.9} & 61.7 & \underline{68.9} & \underline{65.3} & 50.8 & 57.8 \\
    mT-Bi+ & \underline{84.4} & 86.3 & 42.4 & 36.5 & 40.1 & \underline{63.5} & \underline{58.9} & \textbf{86} & 32.5 & \textbf{94.9} & 71.1 & \textbf{65.4} & \textbf{69.2} & \textbf{67.3} & 51.7 & \underline{58.8} \\
    mT-Bi+-CT & \textbf{85.4} & \underline{87.5} & \textbf{43.5} & \textbf{37.8} & \textbf{43.1} & \textbf{73.3} & \textbf{61.8} & \underline{85.7} & \textbf{38.3} & 94.5 & \textbf{72.8} & \underline{63.1} & 63.8 & 63.5 & \textbf{61.3} & \textbf{61.6} \\
    \bottomrule
    \end{tabular}}
\caption{P3 bilingual instruction tuning performance breakdown. The best comparable performances are \textbf{bolded} and second best \underline{underlined}. }
\label{table:p3_bilingual_generalization}
\end{table*}

\subsection{Template Alignment Performance Breakdown}
\label{sec:instruction_alignment_breakdown}
Table \ref{table:instruction_alignment_breakdown} breaks down the scores of bilingual instruction tuning performance of KORANI and P3 respectively. 

\begin{table*}[h!]
    \resizebox{\textwidth}{!}{\begin{tabular}{cccccccccccccccccc}
    \toprule
    \multirow{2}{*}{\textbf{Model}}  & \multicolumn{4}{c}{Sentiment} & \multirow{2}{*}{\textbf{Avg.}} & \multicolumn{5}{c}{Summarization} & \multirow{2}{*}{\textbf{Avg.}} & \multicolumn{5}{c}{Multiple-Choice QA} & \multirow{2}{*}{\textbf{Avg.}}
    \\ \cmidrule(lr){2-5} \cmidrule(lr){7-11} \cmidrule(lr){13-17} & \textbf{Emot.} & \textbf{Rot. Tom.} & \textbf{Ama.} & \textbf{IMDB} & & \textbf{Mul.News} & \textbf{Sam.} & \textbf{CNN.} & \textbf{XSum} & \textbf{Giga.} & & \textbf{Dream} & \textbf{Mul.RC} & \textbf{PiQA} & \textbf{QASC} & \textbf{RACE} & \\
    \midrule
    mT-Ko & 35.8 & 67.7 & \underline{84.9} & 75.7 & 66 & \textbf{7.8} & 11 & \textbf{15.1} & \textbf{11.9} & \textbf{22.9} & \textbf{13.7} & \underline{59.1} & \underline{65.3} & \underline{57.2} & 74.6 & 44.2 & 60.1 \\
    mT-Ko-Trans & \underline{46.4} & \underline{70.7} & \textbf{91.2} & \underline{88.1} & \underline{74.1} & 3.7 & \underline{20} & \underline{13.6} & 9.9 & \underline{19.1} & \underline{13.3} & \underline{59.1} & \textbf{79.8} & 55.8 & \underline{85.7} & \underline{48.8} & \underline{65.8} \\
    mT-Ko-CI & \textbf{49.4} & \textbf{77.8} & \textbf{91.2} & \textbf{89.6} & \textbf{77} & \underline{6.7} & \textbf{20.2} & 13.1 & \underline{10.3} & 18.3 & \textbf{13.7} & \textbf{69} & \textbf{79.8} & \textbf{58.2} & \textbf{91} & \textbf{54.6} & \textbf{70.5} \\
    \midrule
    \bottomrule
    \end{tabular}}
    \resizebox{\textwidth}{!}{\begin{tabular}{ccccccccccccccccc}
    \toprule
    \multirow{2}{*}{\textbf{Model}} & \multicolumn{6}{c}{Natural Language Inference} & \multirow{2}{*}{\textbf{Avg.}} & \multicolumn{3}{c}{Sentence Completion} & \multirow{2}{*}{\textbf{Avg.}} & \multicolumn{2}{c}{Coref. Resol.} & \multirow{2}{*}{\textbf{Avg.}} & WSD & \multirow{2}{*}{\textbf{Total Avg.}}
    \\ \cmidrule(lr){2-7} \cmidrule(lr){9-11} \cmidrule(lr){13-14} \cmidrule(lr){16-16} & \textbf{RTE} & \textbf{CB} & \textbf{AN. R1} & \textbf{AN. R2} & \textbf{AN. R3} & \textbf{EsNLI} & & \textbf{COPA} & \textbf{Hellasw.} & \textbf{StoryC.} & & \textbf{Winogr.} & \textbf{WSC} & & \textbf{WiC} &  \\
    \midrule
    mT-Ko & \underline{65.3} & 75 & \textbf{36.2} & \underline{34} & \textbf{37.5} & 37.5 & 47.6 & \underline{63.7} & \underline{28.9} & 59.1 & 50.6 & \textbf{51.9} & 43.3 & 47.6 & 50.8 & 48.1 \\
    mT-Ko-CI-Trans & \textbf{79.5} & \underline{78} & 31.1 & \textbf{35} & 32.5 & \underline{52} & \underline{51.4} & \textbf{70.3} & 27.7 & \underline{64.1} & \underline{54} & 50.3 & \underline{47.5} & \underline{48.9} & \underline{52.3} & \underline{51.4} \\
    mT-Ko-CI & \textbf{79.5} & \textbf{79.2} & \underline{33.6} & 33.8 & \underline{33.0} & \textbf{53.3} & \textbf{52.2} & 63.3 & \textbf{35.4} & \textbf{67.3} & \textbf{55.3} & \underline{51.6} & \textbf{61.9} & \textbf{56.8} & \textbf{53.3} & \textbf{54.1} \\
    \bottomrule
    \end{tabular}}
\caption{P3 zero-shot cross-lingual generalization performance breakdown. The best comparable performances are \textbf{bolded} and second best \underline{underlined}. }
\label{table:instruction_alignment_breakdown}
\end{table*}

\subsection{Scale up.}
\label{sec:scale_up_breakdown}
\begin{figure*}[ht!]
\centering
    \includegraphics[width=1\textwidth]{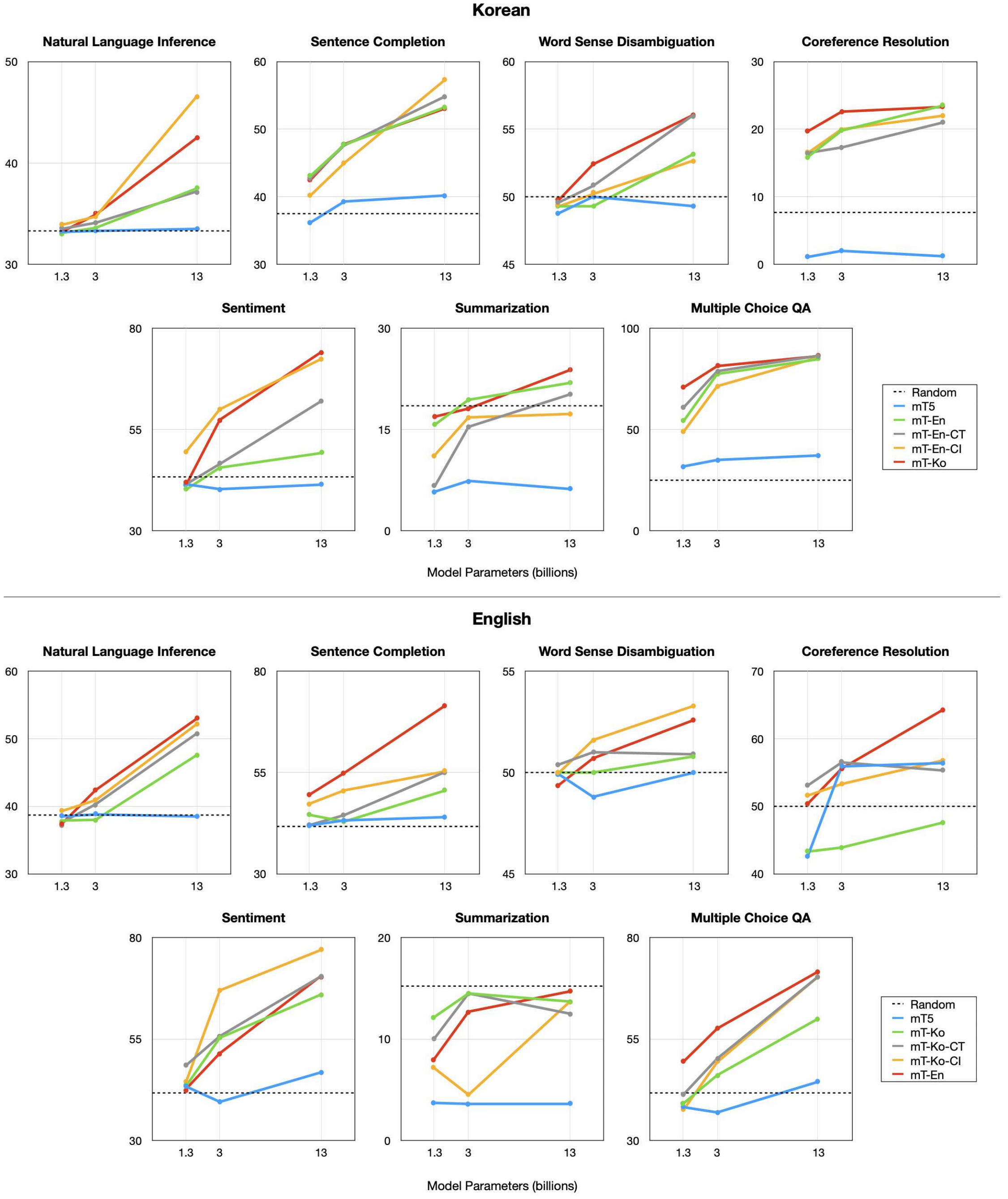}
\caption{Model performance vs. size. performance breakdown. The random line represents the average score random choice in the options list for classification tasks, and the ROUGE-L score of a copy of input for generation tasks.}
\label{fig:scale_up}
\end{figure*}

Figure \ref{fig:scale_up} breaks down the scores average of each task by scaling up from 1.3b to 13b. 

\end{document}